\documentclass[12pt,journal,compsoc,onecolumn]{IEEEtran}
 \usepackage{amsmath,graphicx}
\usepackage{color}
 \usepackage{amssymb}
\usepackage{amsfonts}
\usepackage{dsfont}
\usepackage{epsfig}
 \usepackage{times}
\usepackage{graphicx}
\usepackage{adjustbox}
\usepackage{slashbox}
\usepackage{nicefrac}       
\usepackage{microtype}       
\usepackage{graphicx}
\usepackage{multirow}

 \usepackage{pifont}
\usepackage{lipsum}

 \usepackage{bbm}
 \usepackage{algorithmic}
\usepackage[noadjust]{cite}
 
\newtheorem{proposition}{Proposition}

\ifCLASSINFOpdf
\else
\fi
\ifCLASSOPTIONcompsoc
 \usepackage[font=normalsize,labelfont=sf,textfont=sf]{subfig}
\else
 \usepackage[font=footnotesize]{subfig}
\fi
\ifCLASSOPTIONcompsoc
 \usepackage[font=normalsize,labelfont=sf,textfont=sf]{subfig}
\else
 \usepackage[font=footnotesize]{subfig}
\fi

\usepackage[ruled,vlined]{algorithm2e}

\def\x{{\mathbf x}}

\def\Y{{\cal Y}}
\def\S{{\cal I}}
\def\C{{\cal C}}
\def\x{{\bf x}}
\def\y{{\bf y}}
\def\D{{\cal D}}
\def\CC{{\cal C}}
\def\C{{\bf C}}
\def\F{{\bf F}}
\def\DD{{\bf D}}
\def\tr{{\bf tr}}
\def\1{{\bf 1}}

\title{Active learning for interactive satellite image change detection} 

\author{Hichem Sahbi$^{1}$  \ \ \ \ \ \ \ \ \ \  Sebastien Deschamps$^{1,2}$  \ \ \ \ \ \ \ \ \ \ Andrei Stoian$^{2}$ \\  $ $ \\
$^{1}$Sorbonne University, UPMC, CNRS, LIP6, France \\
$^{2}$Theresis Thales, France}

\begin{document}

\maketitle

\begin{abstract}
We introduce in this paper a novel active learning algorithm for satellite image change detection. The proposed solution is interactive and based on a question \& answer model, which asks an oracle (annotator) the most informative questions about the relevance of sampled satellite image pairs, and according to the oracle's responses, updates a decision function iteratively. We investigate a novel framework which models the probability that samples are relevant; this probability is obtained by minimizing an objective function capturing representativity, diversity and ambiguity. Only data with a high probability according to these criteria are selected and displayed to the oracle for further annotation.  Extensive experiments on the task of satellite image change detection after natural hazards (namely tornadoes)  show the relevance of the proposed method against the related work. \end{abstract}

{\normalsize {\bf keywords:} Frugal learning, interactive satellite image change detection.}

 \section{Introduction} 
 Satellite image change detection is the process of  identifying occurrences of targeted changes in a scene, at a given instant $t_1$, w.r.t. the same scene acquired at an earlier instant  $t_0$. One of the major applications of  change detection is  damage assessment after natural hazards (e.g., tornadoes, earth-quakes, etc.)  in the purpose of  prioritizing disaster response accordingly.  This task consists in  finding relevant changes (such as building destruction,  damage of transportation routes and infrastructures, etc.) while discarding irrelevant ones (due to weather conditions, occlusions, sensor artefact and alignment errors,  as well as effects due to oscillating objects, trees, etc.). \\
 \indent  Early   change detection solutions  were initially based on simple comparisons of multi-temporal signals, via image differences and thresholding, using vegetation indices, principal component and change vector analyses (see \cite{ref7,ref9,ref11,ref13,ref4,ref5} and references therein). These methods also rely on a preliminary pre-processing step that attenuates the effects of irrelevant changes, by finding parameters of sensors for registration as well as correcting radiometric effects, occlusions and shadows (see for instance \cite{ref14,ref2272,ref15,ref17,ref20,ref227}). Other methods either ignore irrelevant changes or consider them as a part of appearance model design and are able to detect relevant changes while being resilient to irrelevant ones \cite{ref21,ref25,ref26,ref27,ref28}. In particular, machine learning  methods are promising~\cite{ref00003,ref00002}, but their success is highly dependent on the availability of large collections of  hand-labeled training data \cite{ref00001,refff3000,refff3001,refff3002}.   Indeed,  these approaches are limited by the huge  scene variability and the lack of labeled training data in order to characterize rare relevant changes and abundant irrelevant ones.  Besides, even when labeled data are available, these labels could be erroneous and oblivious to the annotator's (user's) subjectivity and intention.\\
 \indent Many existing solutions try to bypass  the aforementioned limitations by making learning frugal and less dependent on large collections of labeled data; this includes zero/few shot \cite{reff45,reff58b,ref1015,ref1016,ref1017,ref1018,ref1019,ref1020,ref1021,ref1022,ref1023,ref1024} and self-supervised  learning \cite{refff2,reff58j,ref1000,ref1001,ref1002,ref1003,ref1004,ref1005,ref1006,ref1007}. However, these solutions are limited as their design is agnostic to the user's subjectivity and intention.  Weakly-supervised \cite{ref1008,ref1009,ref1010,ref1011,ref1012,ref1013,ref1014} and active learning solutions \cite{reff1,reff2,reff16,reff53,reff12,reff74,reff58} are rather more appropriate where users annotate very few samples of the most relevant and irrelevant changes (according to a targeted intention), and a training model updates a decision criterion accordingly.  In this paper, we follow this line and we introduce  a novel  interactive satellite image change detection algorithm which asks the user the most informative questions about the relevance of changes and according to the user's responses updates a change detection criterion.  The new proposed model is probabilistic and assigns for each unlabeled training sample a measure which captures how critical is that sample in order to update the decision criteria. We obtain this measure as the optimum of a constrained objective function mixing representativity, diversity, ambiguity and cardinality. In contrast to related active learning solutions \cite{reff13}, which are basically heuristics, the one proposed in this paper is probabilistic and unifies all the aforementioned terms in a single objective function whose solution models the probability of relevance of samples when learning decision functions.  Extensive experiments conducted on the highly imbalanced task of satellite image change detection  show the effectiveness of the proposed model.  
\def\I{{\cal I}} 
\def\p{{p}}
\def\q{{q}}
\def\x{{\bf x}}  
\def\Y{{\cal  Y}}  
 \section{Proposed method}
Let $\I_r = \{\p_1, \dots , \p_n\}$, $\I_t = \{\q_1, \dots , \q_n\}$  be two satellite images captured at two different instants $t_0$ and $t_1$ with $t_0 < t_1$, $\p_i$, $\q_i \in \mathbb{R}^d$  (here $d = 30 \times 30 \times 3$, see experiments). $\I_r$, $\I_t$, referred to as reference and test images respectively, are assumed registered, i.e., pixels in pairs $\{(p_i, q_i)\}_i$ correspond to the same locations. Now, we define $\I = \{\x_1,\dots, \x_n\}$, with $\x_i = (\p_i, \q_i)$, and $\Y = \{\y_1, \dots, \y_n\}$ the underlying unknown labels. Our goal is to  design a change detection function  $f: {\cal I} \rightarrow \{-1,+1\}$ which predicts the unknown labels in $\{\y_i\}_i$ with $\y_i = +1$ if the test patch $\q_i \in  \I_t$  corresponds to a ``change'' w.r.t.  its reference patch $\p_i \in \I_r$ and $\y_i = -1$ otherwise. As ``changes'' are scarce, it is very reasonable to assume that $|\{\x_i : \y_i = +1\}|\ll |\{\x_i : \y_i = -1\}|$.  Learning $f$ requires a subset of training data annotated by an oracle\footnote{i.e., user or expert.}. As these annotations are usually highly expensive, the design of $f$ should consider {\it as few annotations as possible} while minimizing the generalization error $P(f(X)\neq Y)$. 
\subsection{Interactive  satellite image change detection}
Our design principle is iterative and relies on a question \& answer model, which suggests the most informative display\footnote{subset of images.} to an oracle, collects annotations and updates the decision function $f$ accordingly. Let $\D_t \subset \I$ be a subset of images shown to an oracle at iteration $t$ and let $\Y_t$ be the unknown labels of $\D_t$; in practice $|\D_t|$ is fixed depending on the targeted annotation {\it budget}. We build our learning function $f$ iteratively by asking the oracle ``questions'' about labels in $\D_t$ according to the following steps \\

\noindent {\bf Display zero.} Select a display $\D_0$ including representatives of  $\I$. In practice,  $\D_0$ corresponds to centroids  of a partition $\{h_1,\dots,h_K\}$ of $\I$  obtained with K-means clustering.\\ 
\noindent Run the following steps for $t=0,\dots,T-1$ ($T$ may also depend on a predefined annotation budget):\\
{\bf -Oracle model.} Label display $\D_t$ with an oracle function (denoted $\CC(.)$) and assign $\CC(\D_t)$ to $\Y_t$.\\
 {\bf -Learning model.} Train a decision function $f_t (.)$ on data labeled, so far, $\cup_{k=0}^t (\D_k,\Y_k)$ and use it to predict labels  on test data depending on $f_t (.)$. In practice, we use support vector machines (SVMs) in order to build/update $f_t (.)$.\\
 {\bf -Display model.} Select the next display $\D_{t+1}\subset \S\backslash\cup_{k=0}^t \D_k$ (to show to the oracle) that minimizes $P(f_{t+1}(X)\neq Y)$. It is clear that a brute force strategy that (i) considers all the possible displays $\D \subset \S\backslash\cup_{k=0}^t \D_k$, (ii) learns the underlying classifiers $f_{t+1} (.)$ on $\D \cup_{k=0}^t \D_t$ and (iii) evaluates their generalization error  is highly combinatorial and out of reach. Display selection heuristics, related to active learning \cite{reff16aaa}, are usually used instead,  but one should be cautious when using these heuristics since many of them can perform worse than the basic display strategy consisting in choosing data uniformly randomly (see \cite{refff1} and references within).   Our alternative display model proposed in this paper,  combines diversity and ambiguity as well as representativity; diversity aims to select data in order to discover different modes of $f_{t+1} (.)$ while ambiguity seeks to locally refine the decision boundary of $f_{t+1} (.)$. Details about our display model, shown subsequently, constitute the main contribution of this work.
 \def\diag{{\textrm{diag}}}
  \subsection{Proposed display model}
\indent We consider  a probabilistic framework which assigns for each sample $\x_i$ a membership degree $\mu_i$ that measures the probability of $\x_i$ belonging to the subsequent display $\D_{t+1}$; consequently, $\D_{t+1}$ will correspond to the unlabeled $\{\x_i\}_i$ with the highest memberships $\{\mu_i\}_i$. Considering $\mu \in \mathbb{R}^n$ as a vector of these memberships  $\{\mu_i\}_i$, we propose to find  $\mu$ as the minimum of the following constrained minimization problem
{\begin{equation}\label{eq0}
  \begin{array}{l}
    \displaystyle    \min_{\mu \geq 0, \|\mu\|_1=1}  \tr\big(\diag(\mu' [\C \circ \DD])\big) + \alpha [\C' \mu]' \log [\C' \mu] \\
             \ \ \ \ \ \ \  \ \ \ \ \ \ \ \ \ \ \    \ \   + \beta \tr\big(\diag(\mu' [\F \circ \log \F]) \big) + \gamma \mu' \log \mu, 
 \end{array}  
\end{equation}}

\noindent here $\circ$, $'$ are respectively the hadamard product and the matrix transpose, $\|.\|_1$ is the $\ell_1$ norm, $\log$ is applied entry-wise, and $\diag$ maps a vector to a diagonal matrix. In the above objective function, (i) $\DD \in \mathbb{R}^{n \times K}$ and $\DD_{ik}=d_{ik}^2$ is the euclidean distance between $\x_i$ and $k^{\textrm{th}}$ cluster centroid, (ii) $\C \in \mathbb{R}^{n \times K}$ is the indicator matrix with each entry  $\C_{ik}=1$ iff $\x_i$ belongs to the $k^{\textrm{th}}$ cluster ($0$ otherwise), and (iii)  $\F \in \mathbb{R}^{n \times 2}$ is a scoring matrix with $(\F_{i1},\F_{i2})=(\hat{f}_t(\x_i),1-\hat{f}_t(\x_i))$ and $\hat{f}_t \in [0,1]$ being a normalized version of $f_t$. The first term of this objective function (rewritten as $\sum_i \sum_k  1_{\{\x_i \in h_k \}} \mu_i d_{ik}^2$) measures the {\it representativity} of the selected samples in $\D$; in other words, it captures how close is each $\x_i$ w.r.t. the centroid of its cluster, so this term reaches its smallest value when all the selected samples coincide with these centroids. The second term (rewritten as $\sum_{k}   [\sum_{i=1}^n 1_{\{\x_i \in h_k \}} \mu_i] \log [\sum_{i=1}^n 1_{\{\x_i \in h_k \}} \mu_i]$) measures the {\it diversity} of the selected samples as the entropy of the probability distribution of the underlying clusters; this measure is minimized when the selected samples belong to different clusters and vice-versa. The third criterion (equivalent to $\sum_i \sum_c^{nc} \mu_i \F_{ic} \log \F_{ic} $) captures the {\it ambiguity} in $\D$ measured as the entropy of the scoring function; this term reaches it smallest value when  data are evenly scored w.r.t.  different categories. Finally, the fourth term is related to the {\it cardinality} of $\D$, measured by the entropy of the distribution $\mu$; this term also acts as a regularizer and helps obtaining a closed form solution (as also shown subsequently). The impact of these terms is controlled by $\alpha, \beta, \gamma \geq 0$. \\
               We formulate the minimization problem by adding an equality constraint and bounds which ensure a normalization of the relevance values and allow us to consider $\{\mu_i\}_i$  as a probability distribution on $\S$. 
\subsection{Optimization} 
\begin{proposition}
Let $\1_{nc}$, $\1_{K}$ denote vectors of $nc$ and $K$ ones respectively. The optimality conditions of (\ref{eq0}) lead to the solution 
\begin{equation}\label{eq1}
  \mu^{(\tau+1)} :=\displaystyle \frac{\hat{\mu}^{(\tau+1)} }{\|\hat{\mu}^{(\tau+1)}\|_1}, \\
\end{equation} 
with $\hat{\mu}^{(\tau+1)}$ being  
\begin{equation}\label{eq2}
\small 
  \exp\bigg(-\frac{1}{\gamma}[(\DD\circ \C)\1_K + \alpha \C (\log[\C' {\mu}^{(\tau)}]+\1_K)+\beta (\F \ \circ \  \log \F)\1_{nc}] \bigg). 
\end{equation}
                       
\end{proposition} 
In view of space, details of the proof are omitted and  result from the gradient optimality conditions of Eq.~(\ref{eq0}). \\ 
\noindent Considering the above proposition, the optimal solution is obtained iteratively as a fixed point of Eqs (\ref{eq1}) and (\ref{eq2}) with $\hat{\mu}^{(0)}$ initially set to random values. Note that convergence is  observed in practice in few iterations, and the underlying fixed point, denoted as $\tilde{\mu}$, corresponds to the most {\it relevant} samples in the display $\D_{t+1}$ (according to criterion~\ref{eq0}) used to train the subsequent classifier $f_{t+1}$ (see also algorithm~\ref{alg1}).
\begin{algorithm}[!ht]
\KwIn{Images in $\S$, display ${\cal D}_0 \subset {\S}$, budget $T$, $B$.}
\KwOut{$\cup_{t=0}^{T-1} (\D_t,\Y_t)$ and $\{f_t\}_{t}$.}
\BlankLine
\For{$t:=0$ {\bf to} $T-1$}{$\Y_t \leftarrow \CC(\D_t)$  \tcp*[r]{\scriptsize Oracle model}
  $f_{t} \leftarrow \arg\min_{f} P(f(X)\neq Y)$ \tcp*[r]{\scriptsize Learning model (built on top of $\cup_{k=0}^t (\D_k,\Y_k)$)} 
 $\hat{\mu}^{(0)} \leftarrow \textrm{random}$; $\mu^{(0)} \leftarrow \displaystyle \frac{\hat{\mu}^{(0)} }{\|\hat{\mu}^{(0)}\|_1}$; $\tau \leftarrow 0$ \\  
\BlankLine
 \While{($\|\mu^{(\tau+1)}-\mu^{(\tau)}\|_1\geq  \epsilon \ \wedge \ \tau<\textrm{maxiter})$}{
   Set ${\mu}^{(\tau+1)}$ using Eqs.~(\ref{eq1}) and (\ref{eq2}) \tcp*[r]{\scriptsize Display model} 
   $\tau \leftarrow \tau +1$
 }
$\tilde{\mu} \leftarrow \mu^{(\tau)}$ \\ 
$\D_{t+1} \leftarrow \{ \x_i \in \S\backslash \cup_{k=0}^t \D_k: \tilde{\mu}_i \in {\cal L}_B(\tilde{\mu})\}$ \tcp*[r]{\scriptsize ${\cal L}_B(\tilde{\mu})$ being the $B$ largest values of $\tilde{\mu}$}
}
\caption{Display selection mechanism}\label{alg1}
\end{algorithm}
\vspace{-0.08cm}
\section{Experiments} 
We evaluate the performances of our interactive change detection algorithm using a dataset of $2,200$ non-overlapping patch pairs (of $30 \times 30$ pixels in RGB) taken from two registered (reference and test) GeoEye-1 satellite images of $2, 400 \times  1, 652$ pixels with a spatial resolution of 1.65m/pixel. These images correspond to the same area of Jefferson (Alabama) taken respectively in 2010 and in 2011 with many changes due to tornadoes (building destruction, etc.) and no-changes (including irrelevant ones as clouds). The underlying ground truth contains 2,161 negative patch pairs (no-changes and irrelevant ones) and only 39 positive patch pairs (relevant changes), so $< 2\%$ of these patches correspond to relevant changes; half of this set is used to build the display and the learning models and the other half for evaluation. 
Performances are reported using equal error rate (EER) on the eval set of $\I$. EER is the balanced generalization error that equally weights errors in "change" and "no-change" classes. Smaller EER implies better performance.
\subsection{Ablation study}
Initially, we study the impact of each term of our objective function when taken separately and  jointly.  Note that  cardinality term is always kept as it acts as a regularizer that also defines the closed form solution shown earlier. Table~\ref{tab1} shows the impact of each of these terms individually, pairwise and all jointly taken. We observe that representativity+diversity are the most important criteria at the early stages of the iterative change detection process, while the impact of the ambiguity term comes later during the last iterations (when the modes of data are all explored) in order to locally refine change decision functions.
These observations are shown through  EER, and for different sampling rates defined at each iteration $t$ as  $(\sum_{k=0}^{t-1} |\D_k|/(|\I|/2))\times 100$  with $|\D_k|$ set to $16$ in practice and again $|\I|=2,200$. 
 \begin{table}
 \centering
 \resizebox{0.89\columnwidth}{!}{
 \begin{tabular}{c||cccccccccc}
Iter  & 1 &2 & 3& 4& 5& 6& 7& 8& 9 & 10 \\
Samp\%  & 1.45 &2.90 & 4.36& 5.81& 7.27& 8.72& 10.18& 11.63& 13.09 & 14.54 \\
 \hline
 \hline
  rep & 48.05 &  26.21 &  12.72 &   10.48 &    9.88 &   9.70&    8.52 &   8.85&   8.61&   8.82\\
  div&  48.05 &  31.24 &  23.45 &  30.41 &  44.81   &  24.12 &   13.22 &   17.02&    6.88. &   7.98  \\
 amb & 48.05 & 46.68 &   38.73 &   29.91 &   14.74 &   20.11 &   8.33 &   7.41 &   7.37 &   5.53 \\ 
 rep+div & 48.05  &  26.21 &   33.35 &   25.10 &   21.55 &   11.71 &   2.84 &   1.65 &   1.59 &   1.43  \\
 rep+amb &  48.05 &   26.21 &   12.62 &   10.81 &   9.82 &   9.70 &   8.53 &   9.23 &   8.60&   8.82\\ 
 div+amb &   48.05 &  41.69 &   28.82 &  23.08 &   23.41 &   23.42 &   19.82 &   13.10 &   8.16 &   6.97\\ 
all  & 48.05 &   26.21 &  33.35 &   25.52 &   23.70 &  14.59 &   2.74 &   1.54&   1.67   & 1.48  
\end{tabular}}
 \caption{This table shows an ablation study of our display model. Here rep, amb and div stand for representativity, ambiguity and diversity respectively. These results are shown for different iterations $t=0,\dots,T-1$ (Iter) and the underlying sampling rates (Samp) again defined as $(\sum_{k=0}^{t-1} |\D_k|/(|\I|/2))\times 100$.}\label{tab1} 
 \end{table} 
\subsection{Comparison}\label{compare}
\begin{figure}[tbp]
\center
\includegraphics[width=0.59\linewidth]{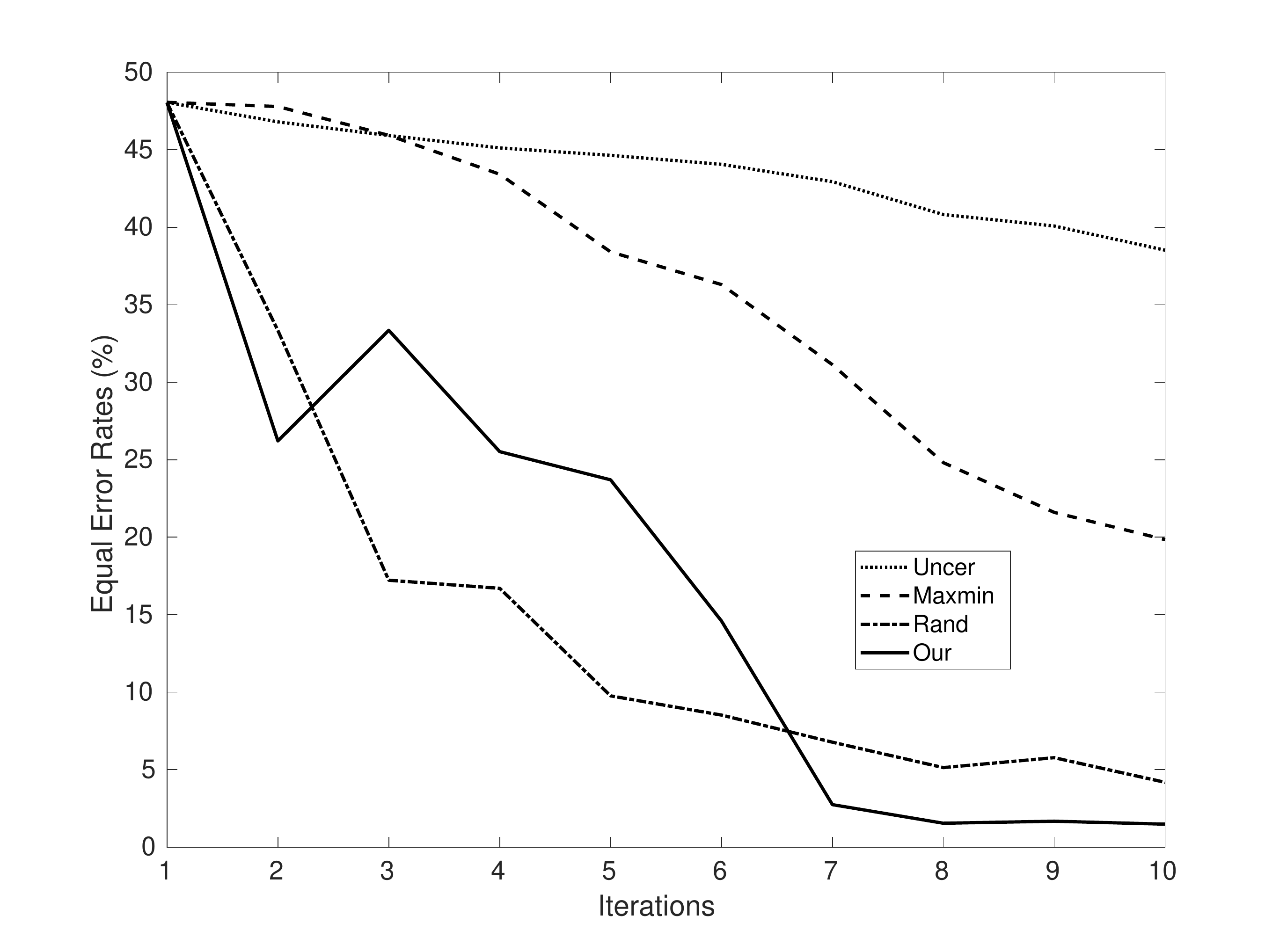}
 \caption{This figure shows a comparison of different sampling strategies w.r.t. different iterations (Iter) and the underlying sampling rates in table~\ref{tab1} (Samp). Here Uncer and Rand stand for uncertainty and random sampling respectively. Note that fully-supervised learning achieves an EER of $0.94 \%$. See again section~\ref{compare} for more details.}\label{tab2}\end{figure}

In order to further investigate the relevance of our method, we compare our display model against three different related strategies, namely {\it maxmin, uncertainty} as well as {\it random sampling}. Maxmin consists in sampling a display  $\D_{t+1}$  greedily at each iteration of the interactive process; each data in  $\x_i \in \D_{t+1} \subset \S\backslash \cup_{k=0}^t \D_k$ is chosen in order to {\it maximize its minimum distance w.r.t.  $\cup_{k=0}^t \D_k$}, thereby the display $\D_{t+1}$  will correspond to the most distinct samples. Uncertainty sampling consists in choosing the display whose unlabeled samples are the most ambiguous (i.e., whose  SVM scores are the closest to zero). Random consists in picking randomly samples from the pool of unlabeled training data. We also compare our method with fully-supervised learning where a monolithic classifier is trained using 100\% of the ground truth of training data.\\
Figure~\ref{tab2} shows EER w.r.t different iterations (and the underlying sampling rates in table~\ref{tab1}). We observe the positive impact of our display model compared to the other strategies. As the task is highly imbalanced, all these comparative strategies are unable to spot the rare class (changes) sufficiently;  maxmin and random make it possible to capture the diversity of the data without being able to minimize the EER at the latest iterations. Uncertainly refines locally the decision functions but suffers from the lack of diversity. Whereas all these strategies (for $t \leq 1$) have high EERs, interactive change detection (when combined with our display model) rapidly reduces the EER and overtakes all the other strategies, at the end of the iterative process. This comes from the rapid adaptation of decision functions $\{f_t(.)\}_t$ to the oracle when frugally learning from the most relevant samples.

\section{Conclusion} 
We introduce in this paper a novel interactive satellite image change detection algorithm. The proposed method is based on a query \& answer model which suggests the most informative patch pairs to an oracle and according to the responses of the latter updates a decision criterion that captures the oracle's intention. Our proposed display model is probabilistic and obtained by minimizing a constrained  objective function  mixing (i) diversity that explores different modes of data distribution, (ii) representativity which focuses on the most informative samples in these modes and (iii) ambiguity that returns samples with the most ambiguous classifications. We also consider a regularization term that smooths  the learned probabilities and allows  obtaining closed form solutions. Experiments conducted on the highly imbalanced task of satellite image change detection  show the  effectiveness of our proposed frugal learning approach.

\end{document}